\documentclass[lettersize,journal]{IEEEtran}
\usepackage{amsmath,amsfonts,amssymb}
\usepackage{algorithmic}
\usepackage{algorithm}
\usepackage{array}
\usepackage[caption=false,font=normalsize,labelfont=sf,textfont=sf]{subfig}
\usepackage{textcomp}
\usepackage{stfloats}
\usepackage{url}
\usepackage{verbatim}
\usepackage{graphicx}
\usepackage{cite}
\usepackage{svg}
\usepackage{hyperref}   
\usepackage{multirow} 
\usepackage{makecell} 
\usepackage{booktabs} 
\usepackage{amssymb}  
\usepackage{tabularx}
\usepackage[normalem]{ulem}  
\usepackage{xcolor}     
\usepackage{float}

\begin{document}

\title{SAIGFormer: A Spatially-Adaptive Illumination-Guided Network for Low-Light Image Enhancement}

\author{Hanting Li, Fei Zhou, Xin Sun,~\IEEEmembership{Senior Member,~IEEE,} Yang Hua, Jungong Han,~\IEEEmembership{Senior Member,~IEEE,}

Liang-Jie Zhang,~\IEEEmembership{Fellow,~IEEE,}
\thanks{H. Li and X. Sun are with Faculty of Data Science, City University of Macau, 999078, SAR Macao, China. F. Zhou is with College of Oceanography and Space Informatics, China University of Petroleum (East China), Qingdao, China. Y. Hua is with School of Electronics, Electrical Engineering and Computer Science, Queen's University Belfast, BT7 1NN Belfast, U.K. J. Han is with Department of Automation, Tsinghua University, Beijing, China. L.J. Zhang is with College of Computer Science and Software Engineering, Shenzhen University, Shenzhen, China. H. Li and F. Zhou contributed to the manuscript equally.}
\thanks{This work is supported by the Science and Technology Development Fund, Macao SAR No.0006/2024/RIA1 and National Natural Science Foundation of China under Project No.61971388.}}

\markboth{Journal of \LaTeX\ Class Files,~Vol.~14, No.~8, August~2025}%
{Shell \MakeLowercase{\textit{et al.}}: A Sample Article Using IEEEtran.cls for IEEE Journals}

\IEEEpubid{}

\maketitle
\begin{abstract}

Recent Transformer-based low-light enhancement methods have made promising progress in recovering global illumination. However, they still struggle with non-uniform lighting scenarios, such as backlit and shadow, appearing as over-exposure or inadequate brightness restoration. To address this challenge, we present a Spatially-Adaptive Illumination-Guided Transformer (SAIGFormer) framework that enables accurate illumination restoration. Specifically, we propose a dynamic integral image representation to model the spatially-varying illumination, and further construct a novel Spatially-Adaptive Integral Illumination Estimator SAI²E. Moreover, we introduce an Illumination-Guided Multi-head Self-Attention (IG-MSA) mechanism, which leverages the illumination to calibrate the lightness-relevant features toward visual-pleased illumination enhancement. Extensive experiments on five standard low-light datasets and a cross-domain benchmark (LOL-Blur) demonstrate that our SAIGFormer significantly outperforms state-of-the-art methods in both quantitative and qualitative metrics. In particular, our method achieves superior performance in non-uniform illumination enhancement while exhibiting strong generalization capabilities across multiple datasets. Code is available at \url{https://github.com/LHTcode/SAIGFormer.git}.

\end{abstract}


\section{Introduction}
\label{intro}
\IEEEPARstart{S}{mart} devices have made capturing images ubiquitous in daily life, yet they frequently produce significantly degraded image quality in uncontrolled environments, especially under low-light conditions. Such poor illumination often arises from slow shutter speeds, high ISO noise, and flash artifacts etc. Therefore, low-light image enhancement (LLIE) is critical in many computer vision tasks \cite{li2021low}, such as object detection \cite{yu2021single} and tracking \cite{wang2024multi}. Various of LLIE methods have been developed with traditional \cite{wang2013naturalness, hao2020low} and deep learning technologies \cite{jiang2019learning, zhang2019zero, lim2020dslr} in last decade years. In contrast to traditional methods that lack robustness in diverse and complex environments, deep learning demonstrates superior performance in LLIE \cite{Chen2018Retinex, zhang2019kindling, zhang2021beyond}. And Vision Transformer (ViT)-based methods \cite{Xu_2022_CVPR, retinexformer, yan2025hvi} recently have shown strong potential, owing to their ability to capture long-range dependencies in data. Most end-to-end methods enhance low-light images by learning a mapping relationship from low-light to normal. Such approaches jointly address complex illumination restoration along with coupled degradations such as color distortion, noise, and artifacts. However, the inherent trade-offs between illumination and degradation factors \cite{hao2022decoupled} during the optimization process make it challenging for these models to handle non-uniform lighting conditions, such as backlit regions and shadows, thereby hindering ideal illumination restoration.

\begin{figure}[t]
        \centering
        \includegraphics[width=0.9\linewidth]{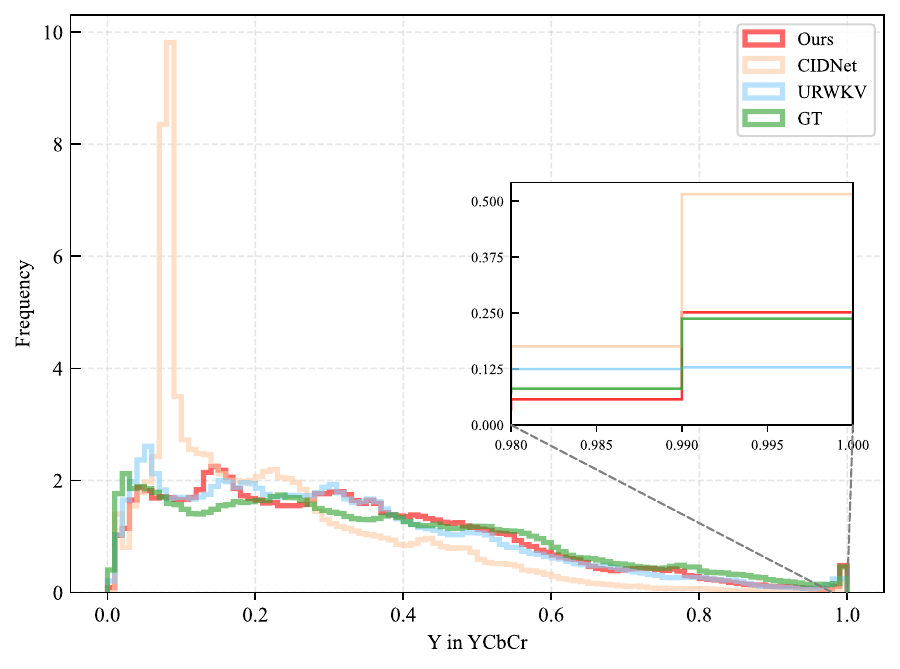}
        \vspace{-10pt}
        \caption{Illustration of the efficient of two representative SOTA methods and ours on illumination restoration. We especially show the Y channels that represent illumination in the YCbCr color space of the image. The estimation results are conducted on 100 images from the LOL-v2-Real, LOL-v2-Syn, and SID datasets via stratified sampling.}
        \label{fig:first-pic}
        \vspace{-18pt}
\end{figure}

To address this challenge, various methods have been proposed. Retinex-based approaches solve this problem by decomposing images into illumination and reflectance components. However, these approaches rely on either redundant network architectures \cite{wu2022uretinex}, precise balancing of multiple loss functions \cite{zhang2019kindling}, or hand-crafted priors with fine parameter tweaking \cite{7782813}, which often leads to poor generalization ability. Some studies overcome such difficulty through multi-stage enhancement frameworks that decouple the illumination restoration process. For example, Hao et al.\cite{hao2022decoupled} proposed a dual-stage network that sequentially performs visibility enhancement followed by fidelity refinement. 
Although multi-stage methods explicitly decouple illumination from entangled degradations, such approaches not only introduce error accumulation across stages but also suffer from ill-defined intermediate objective functions that may deviate the optimization trajectory.

\IEEEpubidadjcol
Although numerous methods have been proposed to decouple illumination from various degradations, existing methods still face significant challenges in recovering non-uniform illumination. For example, one critical limitation is that  current methods struggle to estimate the spatially varying illumination in low-light images accurately, which leads to underfitting of the illumination distribution in the enhanced results. To address the above challenges, we propose one illumination-guided framework called Spatially-Adaptive Illumination Guided Transformer (SAIGFormer). Unlike previous approaches that focus on decoupling the enhancement process or image representations, our method explicitly extracts illumination from the original image to guide the Transformer in learning accurate illumination patterns. Specifically, the essence of our illumination estimator is based on three key insights: \textcolor{brown}{(1) An effective illumination estimation method should be simple and lightweight, avoiding complex network structures and handcrafted constraints.} \textcolor{purple}{(2) It must exhibit strong spatial adaptivity to accurately match the non-uniform and spatially complex illumination distributions.} \textcolor{violet}{(3) It should be located at the early stage, to provide illumination guidance throughout all modules of the framework, for  precise illumination restoration.} Based on these principles, we develop a simple yet efficient Spatially-Adaptive Integral Illumination Estimator (SAI²E), which introduces a dynamic integral image technique to extract the spatially-varying illumination from the original image. Furthermore, we propose one Illumination Guided Multi-head Self-Attention (IG-MSA) module, which integrates the extracted illumination map with channel-wise attention. It essentially calibrates the lightness-relevant features toward visual-pleased illumination enhancement. As shown in Fig. \ref{fig:first-pic}, our method enables accurate estimation of the non-uniform illumination distribution in low-light images. As a result, it outperforms SOTA methods in fitting the illumination distributions, particularly in the poorly and well-illuminated regions. 

Our method achieves promising performance across various benchmarks. In particular, our method surpasses the SOTA methods by 0.33 dB on LOL-v1 dataset, 0.24 dB on the LOL-v2-Syn dataset, 0.23 dB on the SMID dataset, and 0.14 dB on the LOL-Blur dataset, demonstrating both strong performance and remarkable generalization capability. Overall, the major contributions of this work can be summarized as follows:
\begin{itemize}
    \item We propose \textbf{SAIGFormer}, a novel Transformer-based framework for low-light image enhancement, where spatially-adaptive illumination guides the network to accurately enhance complex illumination.
    \item We propose a novel spatially varying illumination estimator, termed \textbf{SAI²E}, which achieves dynamic lighting estimation with $\mathcal{O}(1)$ computational complexity through integral image techniques. To the best of our knowledge, this is the first work to propose dynamic integral image representation in deep learning, especially for low-light image enhancement.
    \item We propose \textbf{IG-MSA} to calibrate channel features, by incorporating the Query component of the attention mechanism guided via illumination, thereby enabling accurate illumination restoration.
    \item Extensive experiments on six datasets demonstrate the superior performance and generalization ability of our method, achieving SOTA results on four datasets and outperforming others on the remaining two.
\end{itemize}

The rest of this paper is organized as follows. Section \ref{sec:related-work} reviews related work in low-light image enhancement, including conventional and deep learning-based approaches, as well as Vision Transformer techniques. Section \ref{sec:methodology} introduces our proposed SAIGFormer framework and its specific modules: SAI²E, SAIGT, IG-MSA, and DG-FFN. Section \ref{sec:experiment} presents both quantitative and qualitative experimental results on various LLIE datasets. Finally, section \ref{sec:conclusion} concludes the proposed method and its contributions.

\section{Related Works}
\label{sec:related-work}
Low-light image enhancement aims to improve the visual perception of images, providing a better visual experience while also benefiting the performance of various high-level vision tasks through enhanced image quality. This section provides a brief introduction to the previous works related to this paper.

In the early studies, research on low-light images enhancement can be broadly categorized into histogram equalization (HE) \cite{6615961}, gamma correction, and Retinex theory \cite{land1977retinex}. Both the original HE and gamma correction methods are regarded as global operations that enhance the overall illumination and visual appeal of an image. However, ignoring the local context often leads to undesirable issues such as noise amplification and color distortion. The Retinex theory assumes that an image can be decomposed into an illumination component and a reflectance component. Therefore, a series of algorithms have been developed by treating the reflectance map as a reasonable approximation of the desired enhanced image. The Retinex theory \cite{land1977retinex} investigated the color constancy property of the human visual system, and argued that the human color perception was not determined by the absolute intensity of light reflected from objects, but rather by their relative reflectance. Some early Retinex-based works \cite{557356, 597272} removed the illumination from the image to obtain the reflectance, which was then treated as the final enhanced result. From then on, researchers focused on designing much reasonable priors and constraints to decompose reflectance and illumination, enhance them, and then recombine them for the enhanced image \cite{7782813, 8304597}. However, methods based on hand-crafted priors and constraints were inherently limited by the model's capacity to accurately decompose reflectance and illumination, making it difficult for them to perform well in challenging and diverse scenarios.

Due to the outstanding performance of deep learning in various computer vision tasks, numerous deep learning related works since 2017 have been conducted \cite{Chen2018Retinex, chen2018learning, zhang2019kindling, Yang_2020_CVPR, Zero-DCE, ZHOU2023109602, jiang2021enlightengan}. Recent year, we have witnessed deep learning methods becoming the mainstream in the field of Low-light image enhancement. Xu et al. \cite{Xu_2022_CVPR} introduced the signal-to-noise ratio (SNR) prior and designed a CNN-Transformer hybrid algorithm based on this prior. Cai et al. \cite{retinexformer} proposed a Retinex theory-based framework called Retinexformer. It first performed an initial light-up of the image and then used the illumination features extracted during this light-up process to guide the Transformer framework in restoring the artifacts. CIDNet \cite{yan2025hvi}, on the other hand, investigated the coupling between image brightness and color in the sRGB space. By using image intensity to represent brightness and decoupling it from color, it designed a Horizontal/Vertical-Intensity (HVI) color space with learnable parameters. Notably, these recently proposed methods all employed Transformer architectures with powerful long-range dependency modeling capabilities. Several approaches (e.g., \cite{zamir2022restormer, retinexformer, ye2023glow, yan2025hvi}) further adopted transformers with transposed attention mechanism (an efficient attention that treats feature maps as tokens with low computational complexity). Other works have explored the characteristics of low-light images in the frequency domain \cite{wang2023fourllie, zou2024wave}. FourLLIE \cite{wang2023fourllie} explored the frequency-domain characteristics of images through Fourier transform and designed a coarse-to-fine two-stage framework, where the first stage enhanced the amplitude spectrum to improve illumination, and the second stage restored image details in the spatial domain. However, it can introduce difficulties in network fitting, solely relying on the amplitude spectrum to restore reasonable illumination. Zou et al. \cite{zou2024wave} proposed Wave-Mamba, which employed wavelet transform to decompose the image into high- and low-frequency components. In the U-Net architecture, the low-frequency components were progressively enhanced along the depth, while the high-frequency signals were propagated and enhanced laterally. However, the enhancement of high-frequency component relied on the low-frequency component, and such decomposition introduced new challenges in domain alignment. In addition, some recent studies in adaptive filtering and dynamic convolution \cite{zhou2023adaptive, sun2021gaussian} exhibited conceptual relevance to our proposed SAI²E module in terms of spatially adaptive processing, although they focus on different tasks.

Despite these progresses, current low-light image enhancement frameworks remain inadequate for accurately estimating non-uniform illumination. In contrast, we propose a novel module that estimates spatially-adaptive illumination component from original images and guides the Transformer through our designed attention mechanism to precisely model illumination features, thereby achieving superior illumination enhancement results.

\section{Methodology}
\label{sec:methodology}

\begin{figure}[t]
        \centering
        \includegraphics[width=1\linewidth]{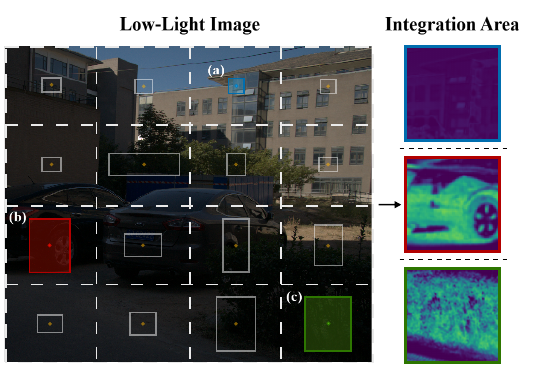}
        \caption{Illustration of the mechanism of the proposed SAI²E, i.e., spatial regions with different lighting conditions should be treated distinctly. Well-illuminated regions (a) require low-pass filters with smaller window sizes for illumination estimation, which appear as small integration areas in the heatmap; whereas poorly illuminated regions (b) and (c) require filters with larger windows, reflected as large integration areas in the heatmap.}
       \label{fig:SAI2E}
        \vspace{-0.3cm}
\end{figure}

\begin{figure*}[!htbp]
        \centering
        \includegraphics[width=0.9\linewidth]{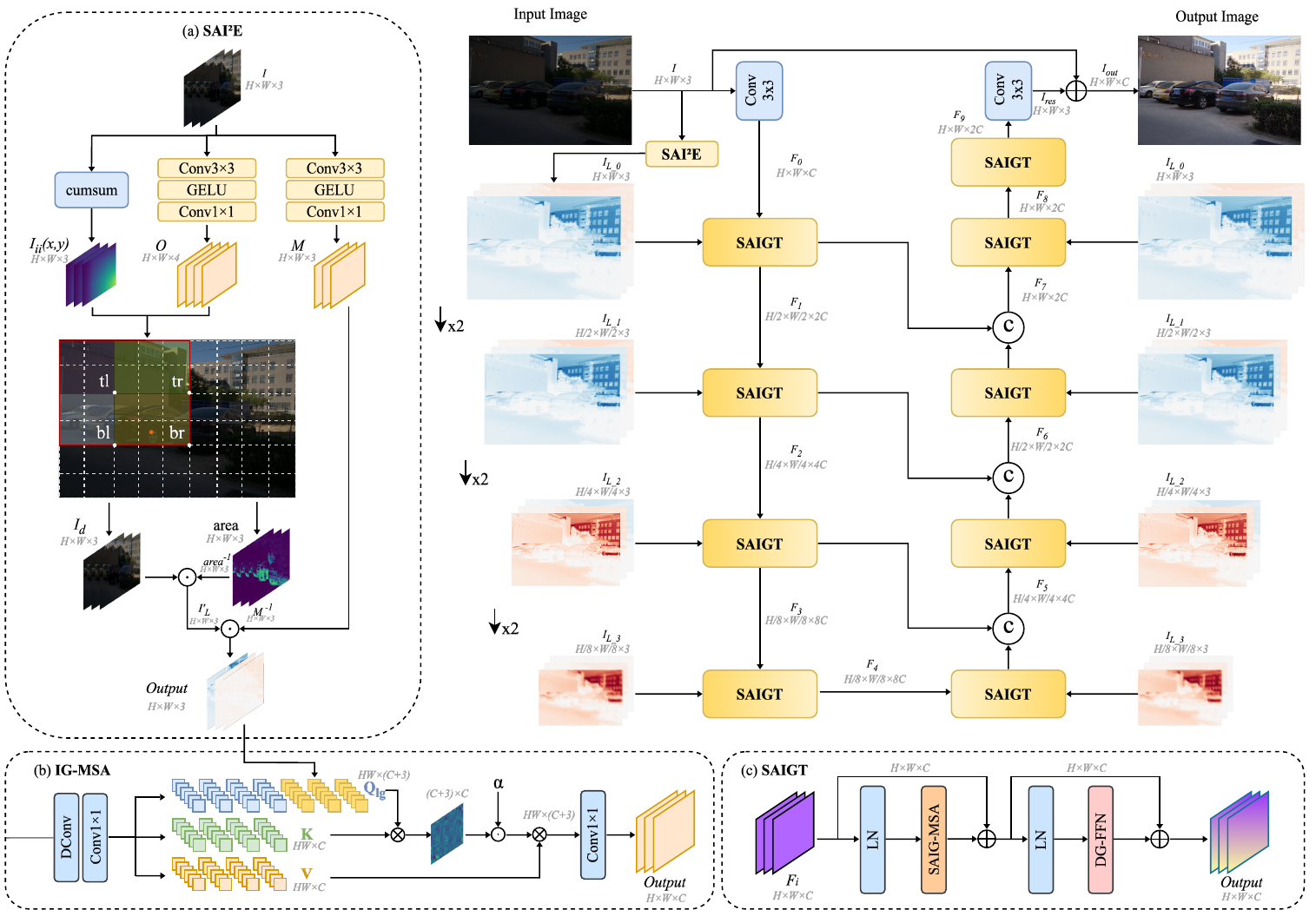}
        \vspace{-5pt}
        \caption{Overview of the SAIGFormer framework. (a) SAI²E adaptively extracts spatially-varying illumination information from the input image. (b) The illumination extracted by SAI²E is directly incorporated into the self-attention computation without being processed by a LayerNorm layer. (c) Each Transformer block is guided by the spatially adaptive illumination, and a Dual-Gated Feed-Forward Network is used to update and retain the learned features.}
        \label{fig:architecture}
\end{figure*}
\subsection{Motivation}
Current approaches \cite{retinexformer, yan2025hvi} typically employ either simplistic Mean-RGB/Max-RGB theory \cite{land1977retinex} or Retinex-based frameworks \cite{wu2022uretinex, zhang2019kindling, 7782813}. Nevertheless, these methods frequently fail to achieve precise illumination restoration. Specifically, Mean/Max-RGB techniques lack spatial adaptability, leading to inaccurate illumination pattern estimation, whereas Retinex-based methods heavily rely on complex network architectures, handcrafted priors, and multi-term loss functions to extract illumination maps.

It is well known that illumination is primarily encoded in the low-frequency components of an image \cite{wang2013naturalness}. In the early Retinex theory-based works \cite{land1977retinex}, Gaussian filtering was employed to enforce smoothness in the estimated illumination map \cite{557356, 597272}, where low-pass filters serve as a viable option for extracting the illumination component. However, illumination in real-world scenes is highly non-uniform and spatially complex. And regions under different lighting conditions exhibit significant variations in noise contamination levels. Therefore, a low-pass filter with a fixed window size cannot consistently capture representative illumination components across all spatial regions. It may mislead the model in learning accurate illumination patterns with such a filter uniformly over the entire image. 

To this end, as illustrates in Fig. \ref{fig:SAI2E}, we propose an efficient illumination estimator, SAI²E, which for the first time introduces a dynamic integral image representation into the low-light image enhancement. Specifically, the integral image \cite{viola2001rapid}, also known as the summed area table, enables fast and parallel computation of the pixel-wise integral over arbitrary rectangular regions of the image. Leveraging this property, we employ tiny sub-networks to adaptively assign low-pass filters with varying window sizes to regions with different illumination conditions in low-light images, thereby enabling accurate illumination estimation for each region. In addition, as emphasized in our third key insight, the illumination estimator should be positioned in the early stages of the network. This is because all components in the backbone contribute to optimizing illumination restoration. Thus, it is essential to provide accurate illumination guidance at each layer for achieving an ideal illumination recovery.

\subsection{Framework Overview}
\noindent As previously mentioned, existing illumination enhancement frameworks struggle to enhance non-uniform illumination effectively. To address this issue, we firstly estimate spatially-adaptive illumination from the original image to guide the Transformer framework in precisely enhance the illumination.

Fig. \ref{fig:architecture}. illustrates the overall framework of the proposed Spatially-Adaptive Illumination Guided Transformer (SAIGFormer). In the shallow layers of the framework, the proposed Spatially-Adaptive Integral Illumination Estimator (SAI²E) estimate illumination from each channel of the input image. The primary architecture of SAIGFormer adopts one U-Net architecture, where each stage is composed of stacked Spatially-Adaptive Illumination Guided Transformer (SAIGT) blocks.
Specifically, SAIGFormer firstly applies the SAI²E and a shallow feature extractor (one $3 \times 3$ convolution) to the original low-light image $I \in \mathbb{R}^{H \times W \times 3}$. It produces a spatially adaptive illumination image $I_{L\_0} \in \mathbb{R}^{H \times W \times 3}$ and initial feature embeddings $F_{0} \in \mathbb{R}^{H \times W \times C}$, where $H$, $W$, and $C$ denote the height, width, and number of channels, respectively. Then, the illumination image $I_{L\_0}$ and the initial embedding $F_0$ are fed into a U-Net architecture composed of multiple SAIGT for feature refinement (e.g., denoising, illumination enhancement etc.) and image enhancement. Finally, a $3 \times 3$ convolution is applied to generate the residual $I_{res} \in \mathbb{R}^{H \times W \times 3}$, which is then element-wise added to the image $I$ to obtain the final result $\hat{I}_{out}$. The SAIGFormer architecture employs a hierarchical pipeline with four spatial resolutions. The feature $F_i$ is downsampled at different stages of the encoder using pixel-unshuffle operations, resulting in resolutions of $\frac{H}{2} \times \frac{W}{2}$, $\frac{H}{4} \times \frac{W}{4}$, and $\frac{H}{8} \times \frac{W}{8}$, and is upsampled to the corresponding resolutions in the decoder using pixel-shuffle operations. Correspondingly, the illumination image $I_{L\_0}$ is downsampled into $I_{L\_1}$, $I_{L\_2}$, and $I_{L\_3}$ at the respective scales using channel-wise depthwise separable convolutions. It is worth noting that the illumination images are not upsampled, but instead shared between the encoder and decoder. Further details of the critical components of our approach are presented below.

\subsection{Spatially-Adaptive Integral Illumination Estimator}
\noindent The proposed Spatially-Adaptive Integral Illumination Estimator (SAI²E) is designed to efficiently and accurately estimate non-uniform illumination from the input image.

The SAI²E consists of four stages: 1) calculation of the integral image, 2) prediction of the integration region and modulation coefficients at each spatial location, 3) computation of the dynamic integral image and estimation of illumination at each spatial location, and 4) modulation of the illumination.

Specifically, we first calculate the integral image of the original image as follows:
\begin{equation}
    I_{ii}(x, y) = \sum_{\substack{x' \leq x \\ y' \leq y}}{I(x', y')},
\end{equation}
where the value at each position $(x, y)$ in the integral image $I_{ii}(x, y)$ represents the sum of all pixel above and to the left of $(x, y)$ in the image $I$.

Then, we design two tiny sub-networks: Offset-Net, which predicts coordinate offsets map $\mathbf{O} \in \mathbb{R}^{H \times W \times 4}$ to determine deformable integral regions, and Modulation-Net, which estimates modulation coefficients map $\mathbf{M} \in \mathbb{R}^{H \times W \times 3}$ to adaptively modulate integral map intensity. The prediction process is formulated as:

\begin{equation}
\begin{aligned}
\mathbf{O} = \text{Conv}_{1 \times 1}(\text{GELU}(\text{Conv}_{3 \times 3}(I))),   \\
\mathbf{M} = \text{Conv}_{1 \times 1}(\text{GELU}(\text{Conv}_{3 \times 3}(I))),
\end{aligned}
\end{equation}

\noindent where the four channels in the offset map $\mathbf{O}$ represent the displacements, denoted as $t$, $l$, $b$, and $r$, of the center coordinate $\mathbf{C}=\{(0,0), (0,1), \ldots (x_c, y_c), \ldots (W, H)\}$ at each spatial location of the top, left, bottom, and right respectively. As we implement random cropping in the training procedure, the sizes of images are not consistent in the training and testing phases. Therefore, we multiply the offset at each spatial location by a scaling factor $N_h = h/H$ and $N_w = w/W$, where $h$ and $w$ are the image size in training, and $H$ and $W$ are the original image dimensions.

The coordinates ($\text{tl}, \text{tr}, \text{bl}, \text{br}$) of the dynamic integration region for each spatial location can be calculated as:
\begin{equation}
\begin{aligned}
\text{x}_{tl} &= x_c - l \cdot N_w, &\quad \text{y}_{tl} &= y_c - t \cdot N_h, \\
\text{x}_{tr} &= x_c + r \cdot N_w, &\quad \text{y}_{tr} &= y_c - t \cdot N_h, \\
\text{x}_{bl} &= x_c - l \cdot N_w, &\quad \text{y}_{bl} &= y_c + b \cdot N_h, \\
\text{x}_{br} &= x_c + r \cdot N_w, &\quad \text{y}_{br} &= y_c + b \cdot N_h,
\end{aligned}
\end{equation}

Then, with the integration region for each spatial location, the dynamic integral image $I_d$ can be calculated as follows:
\begin{equation}
\begin{aligned}
I_d(x,y) = I_{ii}(\text{br}) + I_{ii}(\text{tl}) - I_{ii}(\text{tr}) - I_{ii}(\text{bl}),
\end{aligned}
\end{equation}

Finally, we estimate the illumination at each spatial location and multiply it by the modulation coefficient to obtain the final illumination map $I_L$:

\begin{equation}
\begin{aligned}
&\text{area}(x, y) = (t + b) \cdot (l + r) \cdot \frac{h \times w}{4}, \\
&I'_L(x, y) = \frac{I_d(x,y)}{\text{area}(x, y)}, \\
&I_L(x, y) = I'_L(x, y) \cdot \mathbf{M}^{-1}(x, y).
\end{aligned}
\end{equation}

In summary, we leverage convolutional neural networks in conjunction with the integral image algorithm to adaptively predict low-pass filtering regions with varying window sizes for each spatial location. \textit{Moreover, once the integral image is obtained, each pixel requires only three multiply-add operations with $\mathcal{O}(1)$ complexity, making the proposed SAI²E module highly computationally efficient.}

\subsection{Spatially-Adaptive Illumination Guided Transformer}
\noindent In this section, we design a Transformer block guided by spatially adaptive illumination components of the image. As illustrated in Fig. \ref{fig:architecture}(c), SAIGT consists of two PreLayerNorm (LN), a Illumination Guided Multi-head Self-Attention (IG-MSA) module and a Dual Gated Feed-Forward Network (DG-FFN). The computations within a SAIGT are defined as follows:
\begin{align}
F_i' &= F_i + \text{IG-MSA}(\text{LN}(F_i), I_{L\_i}), \\
F_{i+1} &= F_i' + \text{DG-FFN}(\text{LN}(F_i')).
\end{align}

\noindent{\bf IG-MSA:}
To optimize illumination feature modeling within the Transformer, we propose IG-MSA (Illumination-Guided Multi-head Self-Attention), which integrates illumination into the Query vectors to calibrate channel features, thereby guiding the Transformer toward accurate illumination enhancement.

Specifically, as shown in Fig. \ref{fig:architecture}(b), for the input feature $F_i$ at each layer, we first apply a $1 \times 1$ convolution to aggregate channel-wise information. Then, a $3 \times 3$ depthwise separable convolution is used to encode local spatial information, yielding the query ($\mathbf{Q} \in \mathbb{R}^{H \times W \times C}$), key ($\mathbf{K} \in \mathbb{R}^{H \times W \times C}$), and value ($\mathbf{V} \in \mathbb{R}^{H \times W \times C}$) representations, where $H$, $W$, and $C$ denote the height, width, and number of channels of the feature map, respectively. For simplicity, multi-head formulation is omitted in the notation:
\begin{equation}
\mathbf{Q}, \mathbf{K}, \mathbf{V} = \text{Split}(W_dW_p\text{LN}(F_i)),
\end{equation}
where $W_d$ and $W_p$ denote the $3 \times 3$ depthwise separable convolution and the $1 \times 1$ pointwise convolution, respectively.

To incorporate the illumination $I_L$ while avoiding distribution conflicts with Transformer-encoded features, we propose fusing $I_L$ with the layer-normalized query $\mathbf{Q}$ through a three steps: (1) Adaptive downsampling via a $4 \times 4$ depthwise separable convolution to match the target resolution, (2) Channel alignment using a $1 \times 1$ convolution to harmonize feature statistics, and (3) Channel-wise concatenation to form the illumination-guided query $\mathbf{Q}_{lg} \in \mathbb{R}^{H \times W \times (C+3)}$. This design calibrates the channel features while maintaining compatibility with the Transformer's inherent representations.
\begin{equation}
\begin{split}
I_{L\_i} &= \text{Conv}_{4 \times 4}(I_{L\_i-1}),  \\      
\mathbf{Q}_{lg} &= \text{Concat}(\mathbf{Q},\ W_p I_{L\_i}),
\end{split}
\end{equation}

Subsequently, $\mathbf{Q}_{lg}$ interacts with $\mathbf{K}$ and $\mathbf{V}$ through a channel-wise self-attention mechanism, where $I_L$ participates in computing the affinity between different feature channels. Such an operation makes the illumination map, as a form of feature representation, part of the query vector. It guides the attention mechanism to focus on features that are favorable for estimating and recovering image illumination. Finally, one $1 \times 1$ convolution is applied to aggregate the feature channels weighted by the attention scores:
\begin{equation}
\begin{aligned}
\text{Attention}(\mathbf{Q},& \mathbf{K}, \mathbf{V}, I_{L\_i}) = \mathbf{V} \cdot \text{softmax}\left(\frac{\mathbf{K}^\top \mathbf{Q}_{lg}}{\alpha}\right),      \\
&F_{i}' = W_p\text{Attention}(\mathbf{Q}, \mathbf{K}, \mathbf{V}, I_{L\_i}),
\end{aligned}
\end{equation}

\noindent where $\alpha$ is a learnable scaling parameter, and $W_p$ denotes a $1 \times 1$ convolution layer.

\noindent{\bf Dual Gated Feed-Forward Network:} To further refine the illumination-guided channel features, we introduce a dual-gated mechanism \cite{wang2023ultra} in the feed-forward network. Specifically, two separate $1 \times 1$ convolutions are applied to the input $F_i'$ to aggregate channel-wise information and project the features into a higher-dimensional space. Then, in two parallel paths, GELU and Sigmoid activations are applied respectively, followed by element-wise multiplication for gating. The outputs of the two gated paths are summed and passed through another $1 \times 1$ convolution to adjust the feature distribution and project it back to the original dimension. The DG-FFN is formulated as:
\begin{equation}
\begin{split}
\text{DG-FFN}(F_i') =\ W_p&(\text{GELU}(W_{p1} F_i') \odot W_{p2} F_i' \\ &+ \text{Sigmoid}(W_{p2} F_i') \odot W_{p1} F_i').
\end{split}
\end{equation}

In summary, this paper proposes a novel framework for low-light image enhancement, SAIGFormer. In the design of SAI²E, we introduce, for the first time, a dynamic integral image representation to estimate spatially-adaptive illumination from the original image. The estimated illumination is then used to guide the Transformer's illumination feature modeling, thereby enabling precise illumination restoration.

\begin{table*}[!htbp]
        \renewcommand{\arraystretch}{1.15}
\begin{center}
\caption{Quantitative comparison of different methods on the LOL-v1\cite{Chen2018Retinex} and LOL-v2\cite{yang2021sparse} datasets.}
\label{tab:lol-results}
\begin{tabular}{ p{3.8cm}<{\centering} | p{1.3cm}<{\centering}p{1.3cm}<{\centering} | p{1.3cm}<{\centering}p{1.3cm}<{\centering} | p{1.3cm}<{\centering}p{1.3cm}<{\centering} | p{1.8cm}<{\centering}}
\Xhline{1pt}
\multirow{2}*{\textbf{Methods}} & \multicolumn{2}{c|}{\textbf{LOL-v1}} & \multicolumn{2}{c|}{\textbf{LOL-v2-Real}} & \multicolumn{2}{c|}{\textbf{LOL-v2-Syn}} & \multirow{2}*{Parameter(M)$\downarrow$}\\
~ & PSNR$\uparrow$ & SSIM$\uparrow$ & PSNR$\uparrow$ & SSIM$\uparrow$ & PSNR$\uparrow$ & SSIM$\uparrow$ & ~ \\
\Xhline{1pt}
RetinexNet\cite{Chen2018Retinex} {\tiny (BMVC'18)} & 16.77 & 0.419 & 16.09 & 0.401 & 17.13 & 0.762 & 0.84\\

EnlightenGAN\cite{jiang2021enlightengan} {\tiny (TIP'21)} & 17.48 & 0.652 & 18.64 & 0.677 & 16.57 & 0.734 & 114.35 \\

UFormer\cite{wang2022uformer} {\tiny (CVPR'22)} & 19.61 & 0.755 & 19.41 & 0.657 & 19.66 & 0.871 & 5.29    \\

Restormer\cite{zamir2022restormer} {\tiny (CVPR'22)} & 22.43 & 0.823 &  19.94 & 0.827 & 21.41 & 0.830 & 26.13      \\

MIRNet\cite{zamir2022learning} {\tiny (TPAMI'22)} & 24.14 & 0.830 & 20.02 & 0.820 & 21.94 & 0.876 & 31.76 \\

LLFlow\cite{Wang_Wan_Yang_Li_Chau_Kot_2022} {\tiny (AAAI'22)} & 21.14 & 0.854 & 17.43 & 0.831 & 24.80 & 0.919 & 17.42 \\

SNR-Net\cite{Xu_2022_CVPR} {\tiny (CVPR'22)} & \underline{24.61} & 0.842 & 21.48 & 0.849 & 24.14 & 0.928 & 4.01       \\

LLFormer\cite{wang2023ultra} {\tiny (AAAI'23)} & 23.65 & 0.82 & 20.06 & 0.792 & 24.04 & 0.909 & 24.55     \\

Retinexformer\cite{retinexformer} {\tiny (ICCV'23)} & 23.93 & 0.831 & 22.80 & 0.840 & 25.67 & 0.930 & 1.61  \\

FourLLIE\cite{wang2023fourllie} {\tiny (MM'23)} & - & - & 22.34 & 0.846 & 24.65 & 0.919 & 0.12    \\

GSAD\cite{hou2023global} {\tiny (NeurIPS'23)} & 22.56 & 0.849 & 20.15 & 0.845 & 24.47 & 0.928 & 17.36 \\

RetinexMamba\cite{bai2025retinexmamba} {\tiny (ICONIP'24)} & 24.02 & 0.827 & 22.45 & 0.844 & 25.88 & 0.935 & 24.1 \\

CIDNet\cite{yan2025hvi} {\tiny (CVPR'25)} & 23.50 & \textbf{0.870} & \textbf{23.90} & 0.871 & 25.70 & 0.942 & 1.88\\

URWKV\cite{xu2025urwkv} {\tiny (CVPR'25)} & - & - & 23.11 & \textbf{0.874} & \underline{26.36} & \underline{0.944} & 18.34 \\

\Xhline{1pt}
\textbf{SAIGFormer(Ours)} & \textbf{24.94} & \underline{0.863} & \underline{23.84} & \underline{0.873} & \textbf{26.60} & \textbf{0.946} & 12.35 \\
\Xhline{1pt}
\end{tabular}
\end{center}
\end{table*}

\begin{figure*}[!htbp]
        \centering
        \includegraphics[width=\linewidth]{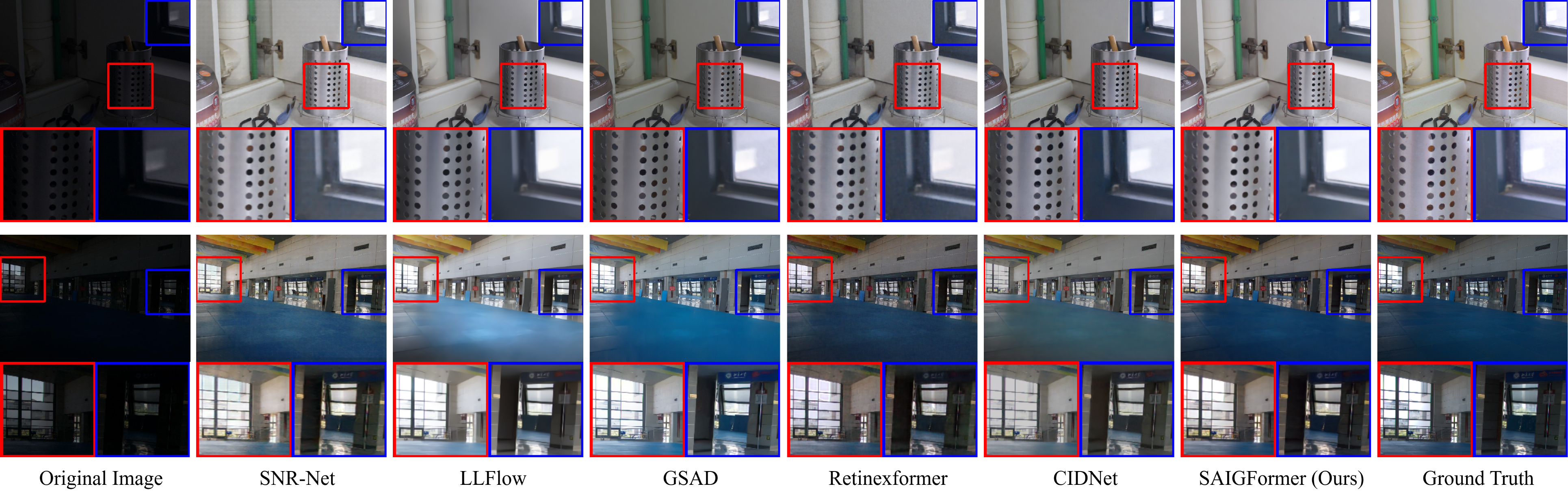}
        \caption{Visual comparison to SOTA models on LOL-v1\cite{Chen2018Retinex} dataset(top) and LOL-v2-real\cite{yang2021sparse} dataset(bottom).}
        \label{fig:lolv1v2results}
\end{figure*}

\section{Experiment}
\label{sec:experiment}
\noindent In section \ref{sec: datasets-and-implementation-details}, we first introduce the dataset setup and implementation details in our experiments. We conduct extensive experiments on multiple datasets to evaluate the performance of our SAIGFormer using standard metrics including PSNR and SSIM\cite{wang2004image}. Then, in Section \ref{sec: compare-with-state-of-the-art-methods}, we present comparisons with SOTA methods along with visual results. In Section \ref{sec: ablation-study}, we provide ablation studies to analyze the effectiveness of different components in SAIGFormer, followed by a discussion and conclusion. 

\subsection{Datasets and Implementation Details}
\label{sec: datasets-and-implementation-details}
\subsubsection{Datasets}
\noindent Our method is evaluated on six benchmarks, including LOL (v1\cite{Chen2018Retinex} and v2-Real and v2-Syn\cite{yang2021sparse}), SID\cite{chen2018learning}, SMID\cite{chen2019seeing}, and LOL-Blur\cite{zhou2022lednet}.


\noindent{\bf LOL v1 and v2:} LOL v1 and v2 are widely used as standard benchmarks in the field of low-light image enhancement. The LOL v1 dataset contains 500 paired images, with 485 used for training and 15 for testing. LOL v2 is an extended version of LOL v1, consisting of two subsets: LOLv2-real and LOLv2-synthetic, which are split into training and testing sets with ratios of 689:100 and 900:100, respectively, following common practice.

\begin{table*}[!htbp]
        \renewcommand{\arraystretch}{1.1}
\caption{Quantitative comparison of different methods on the SID\cite{chen2018learning} dataset.}
\label{tab:sid-results}
\begin{center}
\begin{tabularx}{0.95\linewidth}{X<{\centering\arraybackslash} | X<{\centering\arraybackslash} X<{\centering\arraybackslash} X<{\centering\arraybackslash} X<{\centering\arraybackslash} X<{\centering\arraybackslash} X<{\centering\arraybackslash} X<{\centering\arraybackslash} X<{\centering\arraybackslash}}
\hline
\multirow{2}*{Methods} & SID & RetinexNet & EnlightenGAN  & Uformer & Restormer & MIRNet & SNR-Net & LEDNet \\

~ & \cite{chen2018learning} & \cite{Chen2018Retinex} & \cite{jiang2021enlightengan} & \cite{wang2022uformer} & \cite{zamir2022restormer} & \cite{zamir2022learning} & \cite{Xu_2022_CVPR} & \cite{zhou2022lednet} \\
\hline

PSNR$\uparrow$ & 16.97 & 16.48 & 17.23 & 18.54 & 22.27 & 21.36 & 22.87 & 21.47\\

SSIM$\uparrow$ & 0.591 & 0.578 & 0.543 & 0.577 & 0.649 & 0.632 & 0.625 & 0.638\\

\hline
\hline

\multirow{2}*{Methods} & LLFormer & FourLLIE & Retinexformer & RetinexMamba & MambaIR & URWKV & CIDNet & \textbf{SAIGFormer} \\

~ & \cite{wang2023ultra} & \cite{wang2023fourllie} & \cite{retinexformer} & \cite{bai2025retinexmamba} & \cite{guo2024mambair} & \cite{xu2025urwkv} & \cite{yan2025hvi} & \textbf{(Ours)} \\
\hline

PSNR$\uparrow$ & 22.83 & 18.42 & \textbf{24.44} & 22.45 & 22.02 & 23.11 & 22.90 & \underline{23.50} \\

SSIM$\uparrow$ & 0.656 & 0.513 & \underline{0.680} & 0.656 & 0.658 & 0.673 & 0.638 & \textbf{0.687} \\

\hline
\end{tabularx}
\end{center}
\end{table*}

\begin{figure*}[!htbp]
        \centering
        \includegraphics[width=\linewidth]{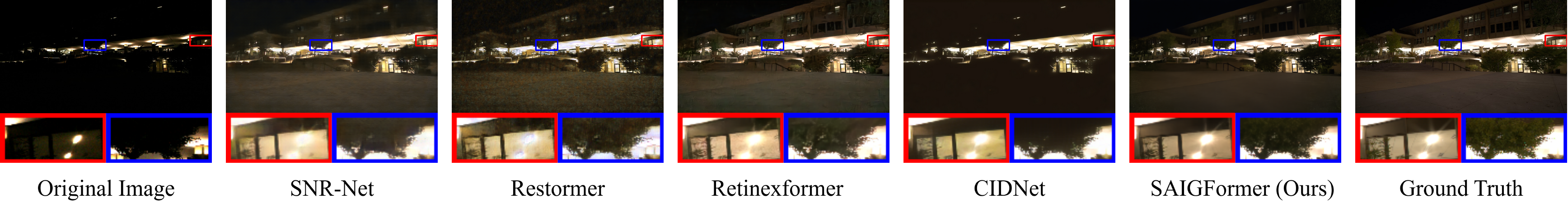}
        \caption{Visual comparison to SOTA models on SID dataset.}
        \label{fig:sidresults}
\end{figure*}

\noindent{\bf SID and SMID:} The SID and SMID datasets are two challenging benchmarks in both the RAW and sRGB domains, with severe noise caused by extremely low-light conditions. In SID and SMID, short- and long-exposure image pairs are treated as low-light and normal-light samples, respectively. The SID dataset contains 2,697 short/long-exposure image pairs and consists of two subsets. As instructed in \cite{chen2018learning}, we apply the same in-camera signal processing pipeline to convert both short- and long-exposure images from RAW to sRGB. We adopt the standard data split using 2,099 images for training and 598 images for testing. The SMID dataset contains a total of 20,809 short-/long-exposure RAW image pairs. Similarly, we convert the RAW data to the sRGB domain for our experiments. We use 15,763 pairs for training and the remaining pairs for testing.

\noindent{\bf LOL-Blur:}
Due to dim environments and the common use of long exposure, images captured under low-light conditions often suffer from both insufficient illumination and motion blur. The LOL-Blur dataset contains images that exhibit both low-light degradation and motion blur, making it a benchmark that presents the dual challenges of low-light enhancement and deblurring. It consists of 12,000 paired low-blur and normal-sharp images, and is split into training and testing sets using a standard 17:3 ratio.

\subsubsection{Implementation Details}
The input image is first embedded into a 32-channel feature map and then fed into the network for enhancement. The numbers of SAIGT Blocks in the encoder, decoder, and refinement stages are set to [4, 6, 6, 8, 6, 6, 4, 4]. Our network is trained for 300k iterations with an initial learning rate set to $2\times10^{-4}$, and a batch size of 8, using the Adam optimizer ($\beta_1 = 0.9$, $\beta_2 = 0.999$). The learning rate is gradually decayed to $1\times10^{-6}$ following a cosine annealing schedule\cite{loshchilov2017sgdr}. During training, we augment the data with random flipping and rotation by 90, 180, and 270 degrees, and randomly crop the input images to a size of $128\times128$. Finally, the training of the network is constrained by a combination of L1 loss and SSIM loss, with the specific form of the SSIM loss defined in Eq. \eqref{eq: ssim}. All experiments are conducted on one single NVIDIA RTX 4090 GPU.
\begin{equation}
\begin{split}
\mathcal{L}_{ssim}(\hat{I}, I_{gt}) = \operatorname{mean}(1 - \mathrm{SSIM}(\hat{I}, I_{gt})).
\label{eq: ssim}
\end{split}
\end{equation}

\subsection{Compare with State-of-the-Art Methods}
\label{sec: compare-with-state-of-the-art-methods}
\noindent
To validate the effectiveness of our method in low-light image enhancement, we compare our approach with SOTA methods on the LOL dataset, LOL-v2-Real, LOL-v2-Syn dataset, SID dataset, SMID dataset, and LOL-Blur datasets. For a fair comparison, we obtain the results of these methods from the publicly available code and pretrained models provided by the respective authors or from their corresponding papers. Tab. \ref{tab:lol-results} presents the performance of our method compared to other approaches on the LOL-v1, LOL-v2-Real, and LOL-v2-Syn datasets, while Tab. \ref{tab:sid-results} and Tab. \ref{tab:smid-results} demonstrate the comparative experimental results on the SID and SMID datasets. Table \ref{tab:lol-blur-results} showcases a series of results on the LOL-Blur dataset; in particular, the results of LEDNet, MIRNet, FourLLIE, LLFormer, Restormer, Retinexformer, GLARE, MambaIR, and URWKV on the LOL-Blur dataset are taken from \cite{xu2025urwkv}.  In all the tables, we use boldface to indicate the best performance and underline to indicate the second-best. 

\begin{table*}[!htbp]
        \renewcommand{\arraystretch}{1.1}
\caption{Quantitative comparison of different methods on the SMID\cite{chen2019seeing} dataset.}
\label{tab:smid-results}
\begin{center}
\begin{tabularx}{0.95\linewidth}{X<{\centering\arraybackslash} | X<{\centering\arraybackslash} X<{\centering\arraybackslash} X<{\centering\arraybackslash} X<{\centering\arraybackslash} X<{\centering\arraybackslash} X<{\centering\arraybackslash} X<{\centering\arraybackslash} X<{\centering\arraybackslash}}
\hline
\multirow{2}*{Methods} & SID & RetinexNet & EnlightenGAN & RUAS & Uformer & Restormer & MIRNet & SNR-Net \\

~ & \cite{chen2018learning} & \cite{Chen2018Retinex} & \cite{jiang2021enlightengan} & \cite{liu2021retinex} & \cite{wang2022uformer} & \cite{zamir2022restormer} & \cite{zamir2022learning} & \cite{Xu_2022_CVPR} \\
\hline

PSNR$\uparrow$ & 24.78 & 22.83 & 22.62 & 25.88 & 27.20 & 26.97 & 26.21 & 28.49\\

SSIM$\uparrow$ & 0.718 & 0.684 & 0.674 & 0.744 & 0.792 & 0.758 & 0.769 & 0.805\\

\hline
\hline

\multirow{2}*{Methods} & LEDNet & LLFormer & FourLLIE & Retinexformer & RetinexMamba & MambaIR & URWKV & \textbf{SAIGFormer} \\

~ & \cite{wang2023ultra} & \cite{wang2023fourllie} & \cite{retinexformer} & \cite{bai2025retinexmamba} & \cite{guo2024mambair} & \cite{xu2025urwkv} & \cite{yan2025hvi} & \textbf{(Ours)} \\
\hline

PSNR$\uparrow$ & 28.42 & 28.42 & 25.64 & 29.15 & 28.62 & 28.41 & \underline{29.44} & \textbf{29.67} \\

SSIM$\uparrow$ & 0.807 & 0.794 & 0.750 & 0.815 & 0.809 & 0.805 & \underline{0.826} & \textbf{0.831} \\

\hline
\end{tabularx}
\end{center}
\end{table*}

\begin{figure*}[!htbp]
        \centering
        \includegraphics[width=\linewidth]{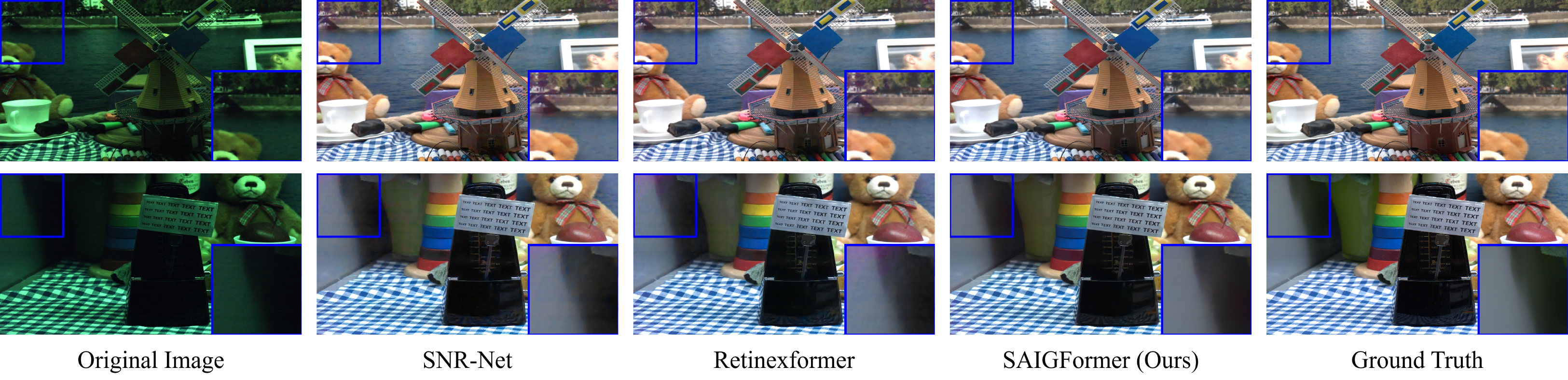}
        \caption{Visual comparison to SOTA models on SMID dataset.}
        \label{fig:smidresults}
\end{figure*}

\noindent{\bf Quantitative Results on LOL-v1 Dataset:}
As shown in Tab. \ref{tab:lol-results}, our method achieves superior performance on the LOL-v1 dataset with notably fewer parameters, outperforming all approaches published in the past three years and establishing new SOTA results. Specifically,  we attain first place in PSNR and SSIM. Notably, our method outperforms the second-best SNR-Net \cite{Xu_2022_CVPR} by 0.33 dB and the third-best MIRNet \cite{zamir2022learning} by 0.8 dB.

\noindent{\bf Quantitative Results on LOL-v2-Real Dataset:}
Our method achieves second best performance in both PSNR and SSIM metrics. Significantly, as shown in Tab. \ref{tab:lol-results}, our method demonstrates superior performance among methods proposed in the past three years, surpassing the third-ranked URWKV \cite{xu2025urwkv} by 0.73 dB and showing only a 0.06 dB gap with the best approach CIDNet \cite{yan2025hvi}.

\noindent{\bf Quantitative Results on LOL-v2-Syn Dataset:}
Our method achieves SOTA performance across all metrics. Specifically, as shown in Tab. \ref{tab:lol-results}, we outperform the second-best URWKV \cite{xu2025urwkv} by 0.24 dB and significantly surpass the third-best RetinexMamba \cite{bai2025retinexmamba} by 0.72 dB. Experiments on the LOL-v1, LOL-v2-Real, and v2-Syn benchmarks demonstrate that our method exhibits robust and stable low-light enhancement performance across diverse and complex scenes.

\noindent{\bf Quantitative Results on SID and SMID Datasets:}
To further validate our model's robust low-light enhancement capability, we conduct comparative experiments on the challenging SID and SMID datasets. Our method achieves the second-best PSNR and the highest SSIM on the SID dataset. Notably, as shown in Tab. \ref{tab:sid-results}, our method closely follows Retinexformer \cite{retinexformer} and achieves the best performance among all methods proposed in the past two years, outperforming the third-ranked URWKV \cite{xu2025urwkv} by 0.39 dB. On the SMID dataset, as shown in Tab. \ref{tab:smid-results}, our method ranks first, outperforming the second-best URWKV \cite{xu2025urwkv} by 0.23 dB and the third-best Retinexformer\cite{retinexformer} by 0.52 dB. Similarly, our method achieves the best performance among all approaches proposed in the past three years.

\begin{table}[!htbp]
        \renewcommand{\arraystretch}{1.2}
\begin{center}
\caption{Quantitative comparison of different methods on the LOL-Blur\cite{zhou2022lednet} dataset.}
\label{tab:lol-blur-results}
\begin{tabularx}{\linewidth}{ X<{\centering\arraybackslash} | X<{\centering\arraybackslash}X<{\centering\arraybackslash}X<{\centering\arraybackslash}X<{\centering\arraybackslash} }
\hline
\multirow{2}*{Methods}  & LEDNet & MIRNet & FourLLIE & LLFormer   \\

~ & \cite{zhou2022lednet} & \cite{zamir2022learning} & \cite{wang2023fourllie} & \cite{wang2023ultra}   \\

\hline
PSNR$\uparrow$ & 26.06 & 23.99 & 19.81 & 24.55   \\

SSIM$\uparrow$ & 0.846 & 0.774 & 0.683 & 0.785   \\
\hline
\hline

\multirow{2}*{Methods} & Restormer & MambaIR & GLARE & Retinexformer   \\

~ & \cite{zamir2022restormer} & \cite{guo2024mambair} & \cite{zhou2024glare} & \cite{retinexformer}    \\

\hline
PSNR$\uparrow$ & 26.38 & 26.28 & 23.26 & 25.25     \\

SSIM$\uparrow$ & 0.860 & 0.848 & 0.690 & 0.821     \\

\hline
\hline

\multirow{2}*{Methods} & PDHAT & URWKV & CIDNet & SAIGFormer   \\

~ & \cite{li2024perceptual} & \cite{xu2025urwkv} & \cite{yan2025hvi} & (Ours)  \\

\hline
PSNR$\uparrow$ & 26.71 & \underline{27.27} & 26.57 & \textbf{27.41}  \\

SSIM$\uparrow$ & 0.885 & \underline{0.890} & \underline{0.890} & \textbf{0.908}  \\

\hline
\end{tabularx}
\end{center}
\end{table}

\begin{figure*}[!htbp]
        \centering
        \includegraphics[width=0.98\linewidth]{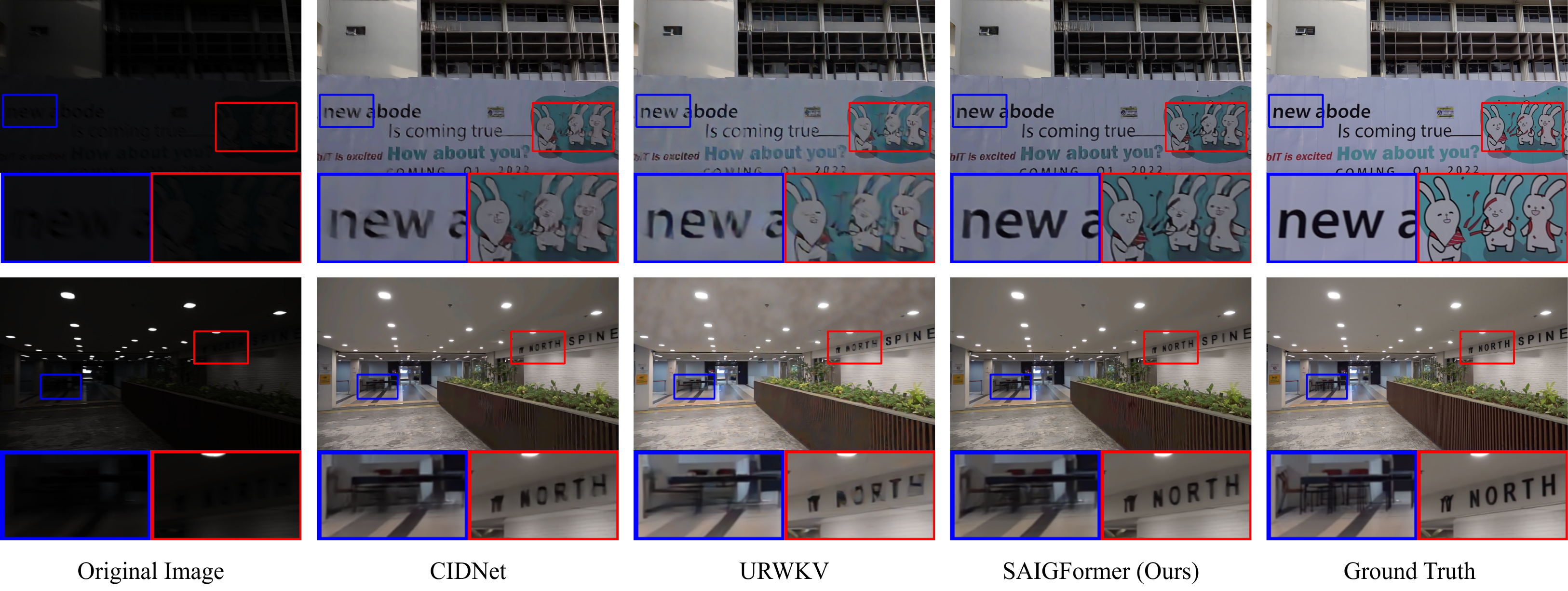}
        \caption{Visual comparison to SOTA models on LOL-Blur dataset.}
        \label{fig:lolblurresults}
        \vspace{-10pt}
\end{figure*}

\noindent{\bf Quantitative Results on LOL-Blur Dataset:}
To demonstrate the potential of our method in addressing various challenge tasks across cross-domain dataset, as well as its strong generalization ability and robustness, we conducted experiments on the LOL-Blur dataset, which involves coupled low-light and motion blur degradations. As shown in Tab. \ref{tab:lol-blur-results}, our method ranks first, outperforming the second-best URWKV \cite{xu2025urwkv} by 0.14 dB and significantly surpasses the third-best PDHAT \cite{li2024perceptual} by 0.7 dB, achieving the best performance among all methods proposed in the past three years. Experiments on the LOL-Blur dataset sufficiently demonstrate the potential of our method for joint low-light enhancement and deblurring, as well as its strong generalization ability and robustness on cross-domain applications. 

\noindent{\bf Visual Results:}
The visual comparisons of SAIGFormer are presented in Fig. \ref{fig:lolv1v2results}, Fig. \ref{fig:sidresults}, Fig. \ref{fig:smidresults} and Fig. \ref{fig:lolblurresults} (zoom in for better viewing). As shown in Fig. \ref{fig:lolv1v2results}, although previous methods have achieved brightness enhancement for low-light images, their lack of accurate illumination guidance for restoration often results in overexposure (e.g., LLFlow, GSAD), underexposure (e.g., CIDNet), or artifacts and noise amplification (e.g., Retinexformer, SNR-Net). In contrast, our method accurately estimates the illumination in the original image and precisely models the illumination features, resulting in enhanced images whose illumination distributions are closest to the ground truth. This is also evident in Fig. \ref{fig:sidresults}, where our method effectively restores illumination and details in backlit and shadowed regions, while avoiding overexposure in already well-lit areas. In contrast, SNR-Net, Restormer, and Retinexformer apply uniform illumination enhancement across all regions, leading to overexposure and artifacts in regions that were already well-lit. On the other hand, while CIDNet does not cause overexposure in well-lit regions, it suffers from noticeable underexposure in backlit and shadowed areas. Fig. \ref{fig:smidresults} further demonstrates the potential of our method in addressing color distortion and noise contamination. Previous methods often suffer from obvious color distortion and insufficient denoising when reconstructing images on the SMID dataset. In contrast, our approach exhibits superior color fidelity, benefiting from the inherent capability of the SAI²E module to capture long-range dependencies, which helps mitigate color artifacts and noise contamination. Moreover, as shown in Fig. \ref{fig:lolblurresults}, our method successfully reconstructs details in extremely dark regions accompanied by severe motion blur. This is attributed to the inherent attention characteristics of the proposed SAI²E, which enhance the model’s ability to capture long-range dependencies within the data. In contrast, methods such as URWKV and CID fail to accurately recover details under such extreme conditions, with URWKV further exhibiting global artifacts.

\subsection{Ablation Study and Analysis}
\label{sec: ablation-study}
\noindent
To demonstrate the effectiveness of each module in our proposed SAIGFormer framework, we conduct extensive ablation studies on the LOL-v2-real dataset.

\noindent{\bf Effectiveness of Proposed Modules:}
The results of the ablation studies for our proposed IG-MSA, SAI²E, and other modules are presented in Tab. \ref{tab: ablation-results}. Here, the baseline experiment 1 refers to one U-shaped network that is solely constructed by stacking Transformers, with its configuration as described in Section \ref{sec: datasets-and-implementation-details}. In experiment 2, the setup preserves the IG-MSA structure but replaces the output of the SAI²E with a commonly used illumination prior (obtained by taking the mean of the input image along the channel dimension) for participating in the computation within the IG-MSA. From experiments 1 and 3, it can be observed that end-to-end deep learning models, without the guidance of spatially adaptive illumination, are unable to accurately enhance the illumination. From experiments 1 and 2, it can be seen that the illumination prior has a certain optimization effect on the learning of end-to-end Transformer-based networks. However, from experiments 2 and 3, it can be observed that illumination without spatial adaptiveness fails to effectively guide the Transformer in modeling illumination features, resulting in unsatisfied illumination enhancement.
\begin{table}[!htbp]
\begin{center}
\caption{Results of the ablation studies.}
\label{tab: ablation-results}
\begin{tabular}{ c | c  c  c | cc }
\Xhline{1pt}
Experiment & baseline & IG-MSA & SAI²E & PSNR & SSIM  \\
\hline
1 & \checkmark & ~ & ~ & 23.01 & 0.867 \\
2 & \checkmark & \checkmark & ~ & 23.22 & 0.871 \\
3 & \checkmark & \checkmark & \checkmark & \textbf{23.84} & \textbf{0.873} \\
\Xhline{1pt}
\end{tabular}
\end{center}
\vspace{-8pt}
\end{table}

\noindent{\bf Different Schemes of SAI²E:}
To validate the design rationale of our proposed SAI²E module, we conduct comparative experiments with alternative configurations. As shown in Tab. \ref{tab: SAI²E-results}, replacing the SAI²E module with non-adaptive avgpool 2x2 leads to decreased PSNR, suggesting that fixed-size low-pass filters, due to their lack of spatial adaptivity, produce inaccurate illumination priors and thereby mislead the Transformer in performing accurate illumination restoration.
\begin{table}[!htbp]
\begin{center}
\caption{different schemes of SAI²E.}
\label{tab: SAI²E-results}
\begin{tabularx}{0.95\linewidth}{ X<{\centering\arraybackslash} | X<{\centering\arraybackslash} X<{\centering\arraybackslash} X<{\centering\arraybackslash} X<{\centering\arraybackslash} }
\Xhline{1pt}
\multirow{2}*{Schemes} &  \multirow{2}*{baseline} & avgpool $2\times2$ & w/o modulation map & \textbf{SAI²E (Ours)} \\
\hline
PSNR$\uparrow$ & 23.01 & 22.83  & 22.95 & \textbf{23.84} \\
SSIM$\uparrow$ & 0.867 & 0.870 & 0.870 & \textbf{0.873} \\
\Xhline{1pt}
\end{tabularx}
\end{center}
\vspace{-8pt}
\end{table}

Furthermore, the PSNR drops when the SAI²E module lacks modulation coefficients, indicating that directly using the SAI²E output for attention computation causes feature distribution discrepancies that mislead Transformer training.

\begin{figure*}[!htbp]
        \centering
        \includegraphics[width=1\linewidth]{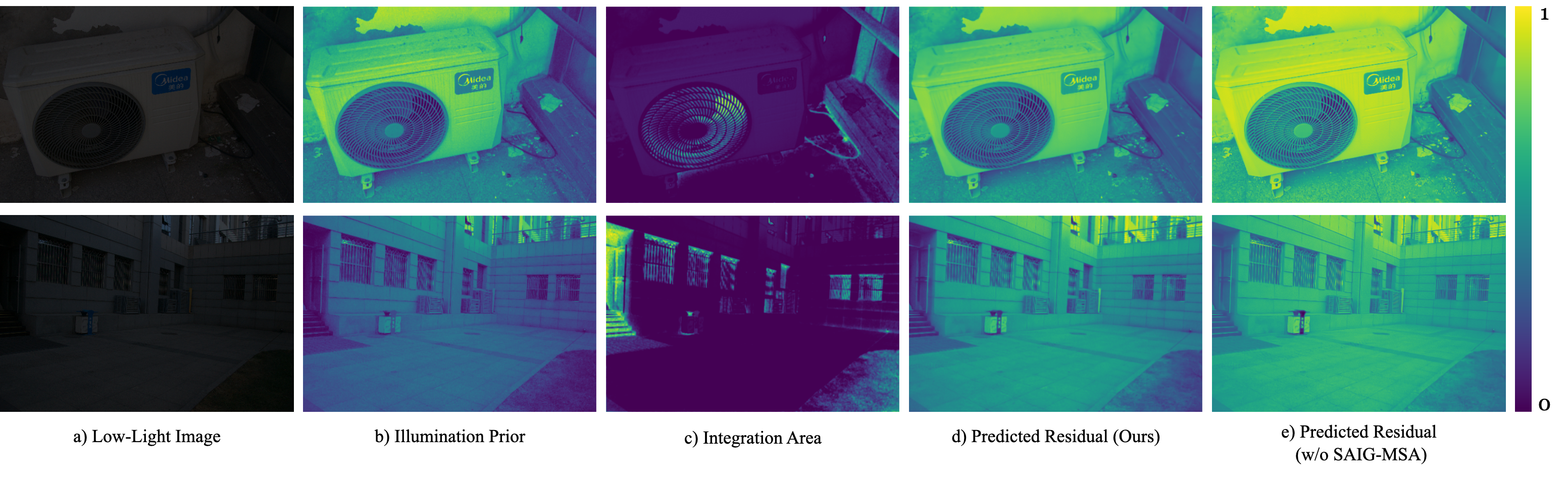}
        \vspace{-25pt}
        \caption{Visual evidence of the effectiveness of the SAI²E module. (b) and (c) illustrate the relationship between the illumination prior of the low-light image (by averaging its channels) and the integration area assigned to each spatial location by the SAI²E module. (d) and (e) present the global residual maps from SAIGFormer and its variant without IG-MSA attention mechanism, respectively. The images reconstructed from these residuals achieve the following performance: for the top row, PSNR = \textbf{27.79}/16.94, SSIM = \textbf{0.940}/0.895; for the bottom row, PSNR = \textbf{30.11}/21.50, SSIM = \textbf{0.909}/0.901, respectively.}
        \label{fig:visualization}
        \vspace{-8pt}
\end{figure*}

\noindent{\bf Visualization Analysis:}
To demonstrate the effectiveness of the proposed SAIGFormer framework, we present several visualization results in Fig. \ref{fig:visualization}. In our heatmap visualizations, we apply min-max normalization to each image to enable comparative analysis of data distributions across different feature maps. As be clearly demonstrated by comparing Fig. \ref{fig:visualization} (b) and (c), our SAI²E module accurately estimates the illumination in original images, adaptively allocating large integration regions to poorly-lit areas while assigning small regions to relatively well-illuminated areas in low-light conditions.

The results in Fig. \ref{fig:visualization} (d) and (e) reveal that, without our proposed IG-MSA module, the end-to-end enhancement framework fails to accurately model illumination features, leading to residual maps with uniform brightening across both dark and bright regions. In contrast, our SAIGFormer benefits from accurate illumination-guided enhancement, producing residual maps that better align with the spatially non-uniform illumination distribution.

\begin{figure}[!h]
        \centering
        \includegraphics[width=0.45\linewidth]{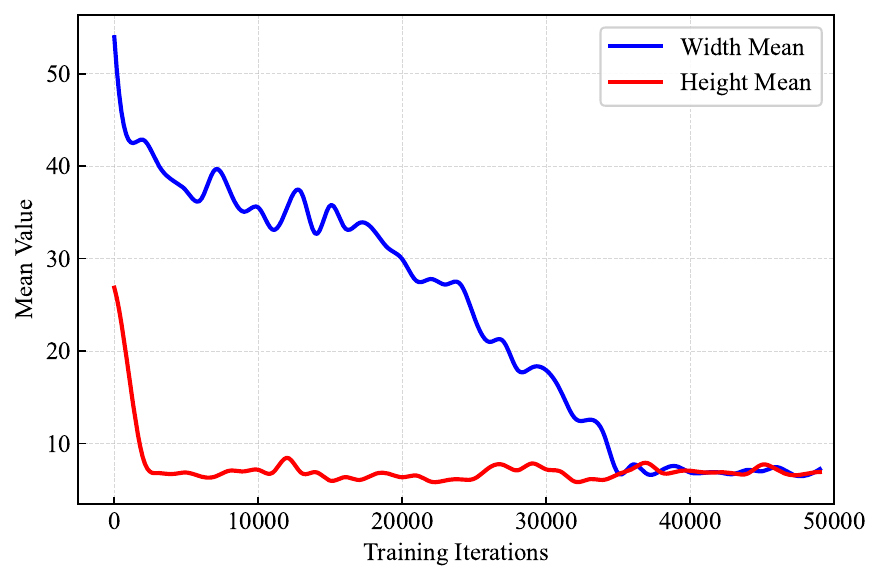}
        \hspace{0.1in}
        \includegraphics[width=0.45\linewidth]{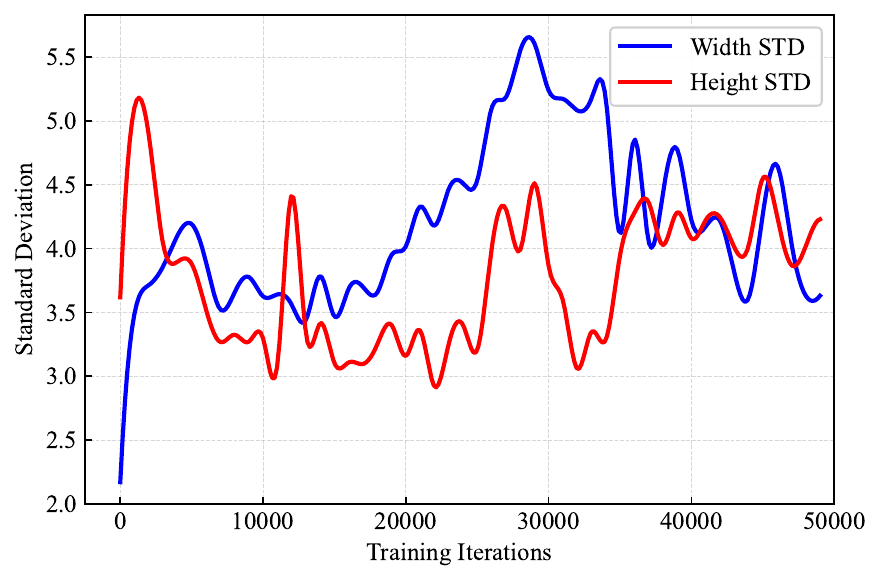}
         \vspace{-8pt}
        \caption{Training dynamics of the SAI²E module. The figure illustrates the distribution of the offset values predicted by the SAI²E module for the same low-light image from the training set at different training stages. The \textbf{left} and \textbf{right} subfigures respectively show the mean and standard deviation of the integral region widths and heights across all spatial locations in the low-light image.}
        \label{fig:mean-std}
         \vspace{-5pt}
\end{figure}
Fig. \ref{fig:mean-std} shows highly active spatial adaptation behavior during training, with significant variations in integration region sizes across different spatial locations of the image. This demonstrates that our designed SAI²E module actively explores satisfied illumination patterns during training to guide the Transformer's illumination reconstruction.

\section{Conclusion}
\label{sec:conclusion}
In this paper, we propose SAIGFormer to address the limitations of existing methods in enhancing illumination under non-uniform lighting conditions. To this end, we introduce a novel illumination estimator, SAI²E, a lightweight algorithm that adaptively matches and estimates complex illumination in the image based on a dynamic integral image representation. Furthermore, to leverage the estimated illumination for enhancement guidance, we design the IG-MSA mechanism, which incorporates illumination into the query vector to calibrate channel features, enabling precise illumination restoration for regions under varying lighting conditions. Owing to these unique designs, our method significantly outperforms state-of-the-art approaches across multiple datasets and demonstrates strong generalization performance on a cross-domain benchmark. In particular, SAI²E offers a novel solution for illumination estimation. Its effectiveness further validates the significant impact of illumination-degradations coupling on the performance of restoration frameworks, and highlights the importance of our three proposed key insights for accurate illumination estimation.
\section*{Acknowledgments}
We would like to express our sincere appreciation to the anonymous reviewers.
\bibliographystyle{IEEEtran}
\bibliography{references}

\newpage

\vfill

\end{document}